\documentclass[10pt,twocolumn,letterpaper]{article}

\usepackage[pagenumbers]{iccv}      

\definecolor{iccvblue}{rgb}{0.21,0.49,0.74}
\usepackage[pagebackref,breaklinks,colorlinks,allcolors=iccvblue]{hyperref}

\def\paperID{6885} 
\def\confName{ICCV}
\def\confYear{2025}

\title{Efficient Density Control for 3D Gaussian Splatting}

\author{ Xiaobin Deng \quad Changyu Diao\footnotemark[1] \quad Min Li \quad Ruohan Yu \quad Duanqing Xu\footnotemark[1]\\
Zhejiang University\\
{\tt\small \{dengxiaobin, dcy, lmin, yuruohan, xdq\}@zju.edu.cn }\\
}

\begin{document}
\twocolumn[{
    \maketitle
	\vspace{-2em}
	\includegraphics[width=1.\linewidth]{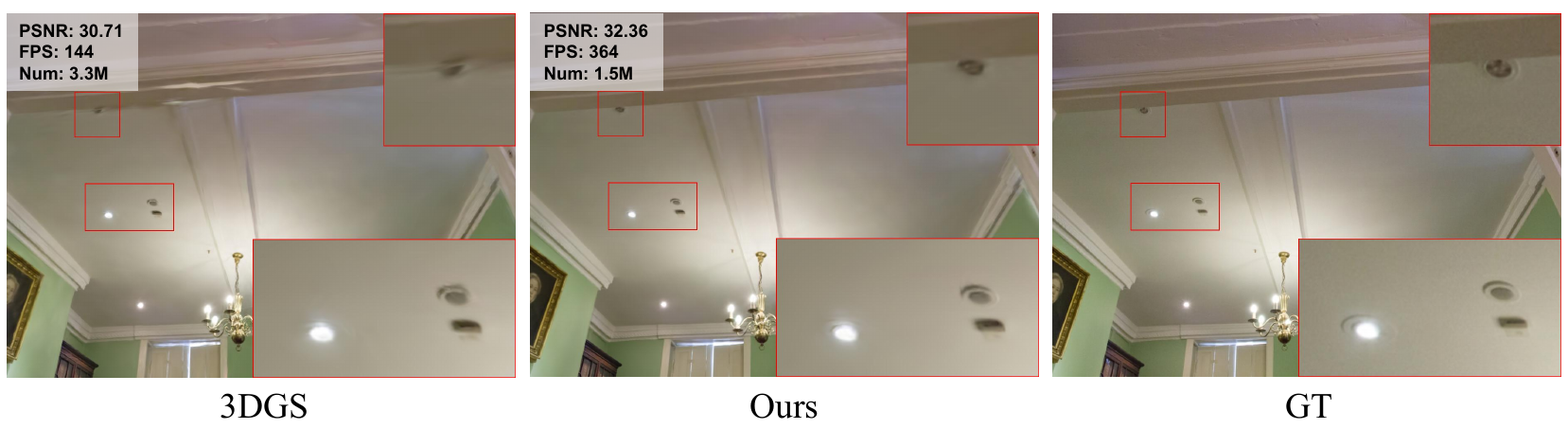}
    \vspace{-20pt}
    \captionof{figure}{By improving adaptive density control of 3DGS \cite{kerbl20233d}, our EDC achieves superior rendering quality while using less Gaussians.
    Benefiting from Long-Aixs Split, we successfully recover details of the ceiling lights and smoke detectors in the drjohnson \cite{hedman2018deep} scene.}
	\vspace{1.3em}
    \label{fig:intro}
}]
{
\renewcommand{\thefootnote}{\fnsymbol{footnote}}
\footnotetext[1]{Corresponding authors.}
}
\begin{abstract}
    3D Gaussian Splatting (3DGS) has demonstrated outstanding performance in novel view synthesis, achieving a balance between rendering quality and real-time performance.
    3DGS employs Adaptive Density Control (ADC) to increase the number of Gaussians. However, the clone and split operations within ADC are not sufficiently efficient, impacting optimization speed and detail recovery.
    Additionally, overfitted Gaussians that affect rendering quality may exist, and the original ADC is unable to remove them.
    To address these issues, we propose two key innovations: (1) Long-Axis Split, which precisely controls the position, shape, and opacity of child Gaussians to minimize the difference before and after splitting.
    (2) Recovery-Aware Pruning, which leverages differences in recovery speed after resetting opacity to prune overfitted Gaussians, thereby improving generalization performance. 
    Experimental results show that our method significantly enhances rendering quality.
\end{abstract}    

Due to resubmission reasons, this version has been abandoned. The improved version is available at: \url{https://xiaobin2001.github.io/improved-gs-web}

\section{Introduction}
\label{sec:intro}

Novel view synthesis (NVS) is a classical problem in computer vision, with widespread applications in virtual reality, cultural heritage preservation, autonomous driving, and other fields.
Neural Radiance Field (NeRF) \cite{mildenhall2021nerf} introduced the use of neural networks to learn the structure and features of a scene, requiring only multi-view 2D images as training data to synthesize high-quality novel views.
However, NeRF suffers from long synthesis times for individual views \cite{muller2022instant, fridovich2022plenoxels}, making real-time rendering challenging.

Recently, 3D Gaussian Splatting (3DGS) \cite{kerbl20233d} has attracted attention due to its explicit representation and real-time rendering performance.
3DGS represents a scene using a large number of 3D Gaussian ellipsoids.
The properties of these Gaussians include position, size, shape, opacity, and color, all of which can be optimized through differentiable rendering.
3DGS generates an initial set of Gaussians from the sparse points obtained through Structure from Motion (SfM) \cite{schonberger2016structure}, and subsequently refines the scene representation by increasing Gaussian density via adaptive density control.

The goal of 3DGS is to achieve high-quality rendering by finely fitting the scene using a large number of 3D Gaussian ellipsoids.
Its training process can be divided into two parts: (1) Increasing the number of Gaussians through adaptive density control. (2) Optimizing the parameters of the Gaussians via backpropagation.
Densification operations are interleaved within the optimization process.
To improve optimization efficiency, we should minimize their negative impact as much as possible.
However, the clone and split operations used in 3DGS fail to meet this requirement.
The clone operation relies on parameter updates from the current iteration to separate it from its clone.
However, this method does not always work effectively.
Since two overlapping Gaussians may receive similar gradients, they cannot be individually optimized to fit the scene accurately.
The split operation generates child's coordinates using probabilistic sampling without altering the shape of the children, leading to a significant discrepancy between the overall shape distribution before and after splitting.
Since the parent's shape is optimized through backpropagation and closely aligns with the scene geometry, the shape changes introduced by splitting negatively impact the optimization process.

\begin{figure}[t]
  \centering
  \includegraphics[width=0.8\linewidth]{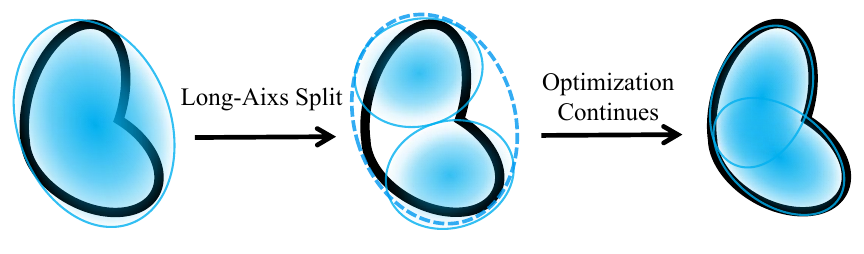}
  \caption{Our proposed Long-Axis Split can minimize the differences before and after splitting, thereby improving the optimization speed post-split.}
  \label{fig:LAS}
  \vspace{-1.0em}
\end{figure}

To minimize the adverse effects of densification operations, we propose Long-Axis Split (see Figure~\ref{fig:LAS}), a more precise alternative, to replace the original clone and split operations.
Long-Axis Split adjusts the position, shape, opacity of the child Gaussians to minimize discrepancies before and after splitting while avoiding Gaussian overlaps.
As Long-Axis Split reduces the negative impact of the densification process, it significantly improves the final rendering quality (see Figure~\ref{fig:intro}).
During the training process, we discover overfitted Gaussians that perform well in certain views but negatively impact rendering quality in other views.
To mitigate the influence of those Gaussians, we propose Recovery-Aware Pruning, which leverages the difference in recovery rates between overfitted and normal Gaussians for pruning.
The newly generated Gaussians contain both overfitted and normally fitted components.
After multiple rounds of Recovery-Aware Pruning, the proportion of overfitted Gaussians among the remaining ones becomes significantly reduced, which notably improves generalization performance.
On the challenging Mip-NeRF 360 dataset, EDC (TamingGS-Abs) achieves a notable improvement in PSNR from $27.48$ to $28.15$ compared to 3DGS, while also reducing the number of Gaussians from $3.3M$ to $2.1M$.

Our contributions are summarized as follows:
\begin{itemize}
  \setlength{\itemsep}{0pt}
  \setlength{\parsep}{0pt}
  \setlength{\parskip}{0pt}
  
      \item We propose Long-Axis Split, a more accurate densification operation, significantly enhancing reconstruction quality.
      \item We introduce Recovery-Aware Pruning, which eliminates overfitted Gaussians that harm generalization performance.
      \item Our method is plug-and-play and brings substantial improvements across multiple 3DGS variants.
      
  \end{itemize}

\section{Related Works}
\label{sec:related}

3D Gaussian Splatting (3DGS) \cite{kerbl20233d} demonstrates outstanding performance in both rendering quality and speed, representing the current state-of-the-art in novel view synthesis.
3DGS has been widely adopted across fields including dynamic scenes \cite{wu20244d, lin2024gaussian}, simultaneous localization and mapping (SLAM) \cite{matsuki2024gaussian, yan2024gs, keetha2024splatam}, 3D content generation \cite{tang2023dreamgaussian, chen2024text, yi2023gaussiandreamer, vilesov2023cg3d}, autonomous driving \cite{zhou2024drivinggaussian}, and high-fidelity human avatars \cite{shao2024splattingavatar, kocabas2024hugs, liu2024humangaussian, moreau2024human}.

Numerous studies have focused on improving 3DGS rendering quality.
For instance, Mip-Splatting \cite{yu2024mip} introduces a 3D smoothing filter and a 2D mipmap filter to eliminate aliasing artifacts present in 3DGS during scaling.
To mitigate the impact of defocus blur on reconstruction quality, Deblurring 3DGS \cite{lee2024deblurring} applies a small multi-layer perceptron (MLP) to the covariance matrix, learning spatially varying blur effects.
GaussianPro \cite{cheng2024gaussianpro} leverages optimized depth and normal maps to guide densification, filling gaps in areas initialized via structure-from-motion (SfM) \cite{schonberger2016structure}.
Spec-Gaussian \cite{yang2024spec} employs anisotropic spherical Gaussian appearance fields for Gaussian color modeling, enhancing 3DGS rendering quality in complex scenes with specular and anisotropic surfaces.
Notably, all these enhancements rely on the original density control and could benefit from our proposed work.

3DGS increases Gaussians in a scene through a basic density control mechanism, yet the optimized scene still exhibits blurred regions that are challenging to refine merely by adding more Gaussians.
MiniGS \cite{fang2024mini} addresses this by generating depth maps for trained scenes to reinitialize the sparse points, and identifies blurred Gaussians with large rendering areas during training, splitting them as needed.
Pixel-GS \cite{zhang2024pixel} addresses blur by using the average gradient weighted by the pixel area covered by Gaussians in each view.
AbsGS \cite{ye2024absgs} and GOF \cite{yu2024gaussian} attribute blur in reconstructions to conflicts in gradient direction across pixels when computing Gaussian coordinate gradients.
This conflict leads to larger Gaussians, which represent blur, receiving insufficient average gradients.
To resolve this, they compute Gaussian coordinate gradients by taking the modulus of pixel coordinate gradients before summing.
TamingGS \cite{mallick2024taming} proposes a densification judgment condition that employs a weighted combination of multiple scores.
The above studies propose improvements for initialization and densification criteria but do not address the densification operation itself.
RevisingGS \cite{rota2024revising} optimizes the opacity bias of Gaussians after cloning.
VCR-GauS \cite{chen2024vcr} replaces the probabilistic sampling used in split operation with a method along the longest axis of the Gaussian, aiming to alleviate surface protrusions caused by the clustering of Gaussians after splitting.
Although these improvements target the densification operation, they do not reduce the negative impact introduced by densification.
We are the first to analyze the adverse effects of densification operations and propose corresponding improvements.

There are also many optimization-based pruning methods. LightGaussian \cite{fan2024lightgaussian} calculates a global importance score for each Gaussian and prunes those with lower scores.
EAGLES \cite{girish2024eagles} uses the average contribution of Gaussians to rendering across all views as the pruning criterion, where the rendering contribution is determined by opacity and rendering order.
MiniGS \cite{fang2024mini} first optimizes the scene structure and then applies binary pruning to retain only the Gaussians that intersect with rays at their first hit point.
Previous pruning methods have focused on significantly reducing the number of Gaussians to lower storage and rendering costs, whereas our Recovery-Aware Pruning focuses on mitigating the effects of overfitting.
\section{Methods}
\label{sec:methods}

\subsection{Preliminaries}
\label{subsec:Preliminaries}

3DGS defines the scene as a set of anisotropic 3D Gaussian primitives:
\begin{equation}
    G(x)=\exp{\left( -\frac{1}{2} (x)^T \Sigma^{-1} (x) \right)},
\end{equation}
where $\Sigma$ is the 3D covariance matrix and $x$ represents the position relative to the Gaussian mean coordinates.
To ensure the semi-definiteness of the covariance matrix, 3DGS reparameterizes it as a combination of a rotation matrix $R$ and a scaling matrix $S$:
\begin{equation}
    \Sigma = R S S^T R^T.
\end{equation}
The scaling matrix $S$ can be represented using a 3D vector $s$, while the rotation matrix $R$ is obtained from the quaternion $q$.
To render an image from a specified viewpoint, the color of each pixel $p$ is obtained by blending $N$ ordered Gaussians $ \{ G_i \mid i = 1, \dots, N \}$ that cover pixel $p$, with the following formula:
\begin{equation}
	\label{eq:front-to-back}
	C = \sum_{i=1}^{N}
	c_{i}\alpha_{i}
	\prod_{j=1}^{i-1}(1-\alpha_{j}),
\end{equation}
where $\alpha_{i}$ is the value obtained by projecting $G_i$ onto $p$ and multiplying by the opacity of $G_i$, while $c_{i}$ represents the color of $G_i$, expressed by SH coefficients.

3DGS initializes the scene using sparse points generated by SfM, and then increases the number and density of Gaussians in the scene through adaptive density control.
Specifically, 3DGS calculates the cumulative average view-space positional gradients of Gaussians every 100 iterations, with each iteration training a single viewpoint.
The formula for calculating the average gradient is as follows:
\begin{equation}
    \frac{\sum_{k=1}^{M^i}{\sqrt{\left( \frac{\partial L_k}{\partial \mu _\mathrm{k,x}^{i}} \right) ^2+\left( \frac{\partial L_k}{\partial \mu _\mathrm{k,y}^{i}} \right) ^2}}}{M^i}>\tau _\mathrm{pos},
\end{equation}
where $M^i$ represents the number of viewpoints in which the Gaussian participates during a cycle, $\tau _\mathrm{pos}$ is the given densification threshold, 
$\frac{\partial L_k}{\partial \mu _\mathrm{k,x}^{i}}$ and $\frac{\partial L_k}{\partial \mu _\mathrm{k,y}^{i}}$ represent the gradients of the Gaussian with respect to the $x$ and $y$ for the current viewpoint, obtained by summing the gradients of each pixel with respect to the coordinates:
\begin{equation}
    \frac{\partial L_k}{\partial \mu _\mathrm{k,x}^{i}} = \sum_{j=1}^m  \frac{\partial L_j}{\partial \mu_{i,x}}, \ \ \
    \frac{\partial L_k}{\partial \mu _\mathrm{k,y}^{i}} = \sum_{j=1}^m  \frac{\partial L_j}{\partial \mu_{i,y}}.
\end{equation}
Gaussians with average gradients exceeding a predefined threshold undergo densification using either clone or split, depending on their size.

\subsection{Analysis of Original ADC}
\label{subsec:limitations}

\textbf{Clone}: Duplicate a small Gaussian with parameters (including position) identical to the parent.
Since the clone operation occurs after the rendering step, the cloned Gaussian does not receive gradients during the current iteration.
In subsequent parameter updates, only the parent Gaussian’s parameters are modified.

\begin{figure}[t]
    \centering
    \includegraphics[width=1.\linewidth]{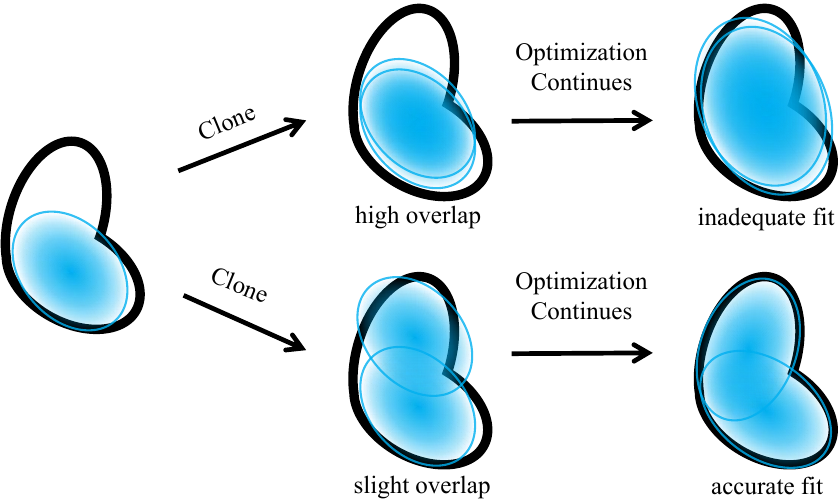}
    \caption{Compared to the normal case (below), two highly overlapping Gaussians after cloning (above) are more difficult to individually optimize for precise scene fitting.}
    \label{fig:clone}
\end{figure}

The degree of overlap between the two Gaussians after cloning depends on the magnitude of parameter changes of the parent Gaussian during the current iteration.
However, not all Gaussians that require cloning exhibit large gradients during the iteration when cloning is performed.
If the parameter update of the parent Gaussian is minimal in the current iteration, severe Gaussian overlap can occur.
Due to receiving similar gradients, highly overlapping Gaussians struggle to be individually optimized in subsequent iterations to fit scene details (see Figure~\ref{fig:clone}).
Experimental analysis is provided in Section~\ref{subsec:reset}.

\begin{figure}[t]
    \centering
    \includegraphics[width=1.\linewidth]{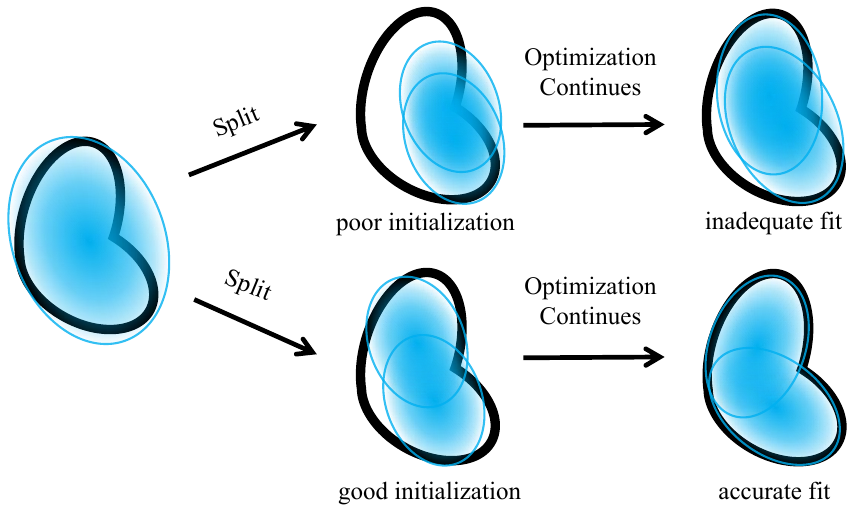}
    \caption{Split uses probabilistic sampling to generate the coordinates of child Gaussians.
    When the initial positions of the two child Gaussians are more reasonable (below), the final fitting result is better.
    Conversely, when the initial positions are less reasonable (above), the final fitting result is poorer.}
    \label{fig:split}
\end{figure}

\textbf{Split}: A large Gaussian is replaced by two smaller Gaussians, which retain the same shape, opacity, and color as the original.
Each smaller Gaussian is scaled down to $1/1.6$ of the parent's size.
The coordinates of the two smaller Gaussians are generated through Gaussian sampling, using the parent’s position and covariance matrix as parameters.

\begin{figure}[t]
    \centering
    \includegraphics[width=0.8\linewidth]{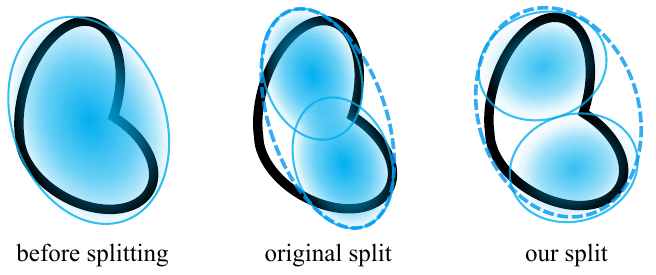}
    \caption{The original split method does not alter the shape of the sub-bodies, resulting in a shape formed by the child Gaussians that differs from the original shape of the parent Gaussian. In contrast, our method shortens the child Gaussians along their longest axis, ensuring that the shape of the covered region before and after the split remains approximately the same, thereby maximizing the densification efficiency.}
    \label{fig:shape}
\end{figure}

During training, the shapes of Gaussians gradually align with the target geometry, including over-reconstructed regions.
Since the two smaller Gaussians maintain the same shape as the original Gaussian, their coverage cannot fully align with the parent’s shape, leading to deviations from the target geometry (see Figure~\ref{fig:shape}). 
This geometric discrepancy slows down convergence during optimization and reduces the final rendering quality.
Additionally, probabilistic sampling introduces extra uncertainty, resulting in greater fluctuations in training outcomes (see Figure~\ref{fig:split}).

\textbf{Reset Opacity}: During densification, an opacity reset operation is performed every $3000$ iterations.
Specifically, the opacity of Gaussians with opacity greater than $0.01$ is reset to $0.01$.

3DGS expects that resetting opacity in conjunction with pruning will help control the number of Gaussians and address floating artifacts.
However, in practice, the combination of these operations has a negligible impact on the number of Gaussians and fails to eliminate floaters.
Still, the process of resetting and subsequently restoring opacity repeatedly optimizes the contribution of different Gaussians in rendering, providing some improvement in reconstruction quality (see Section ~\ref{subsec:reset}).

\subsection{Long-axis Split}
\label{subsec:split}

To better describe the differences between Long-Axis Split and the original split, we will present the explanation in three parts.

\textbf{Position and Shape}: If we consider a Gaussian as a point, the split operation is essentially performed along one dimension.
Among the three axes of a Gaussian, we use its longest axis as the splitting dimension.
The split Gaussians are positioned symmetrically on both sides of the original Gaussian’s longest axis.
To maximize the utilization of the shape information from the original Gaussian, we carefully adjust the shape of the child Gaussians to ensure that the overall shape remains consistent before and after splitting.
Specifically, the radius along the longest axis of each child Gaussian is halved, while the radii along the other axes are shortened to $85\%$ of their original values.
This adjustment also helps mitigate the issue of excessively elongated Gaussians.
Additionally, we set the spacing to the maximum radius of the original Gaussian to prevent overlap (see Figure~\ref{fig:shape}).
\begin{figure}[t]
    \centering
    \begin{minipage}{0.42\linewidth}
        \centering
        \includegraphics[width=\linewidth]{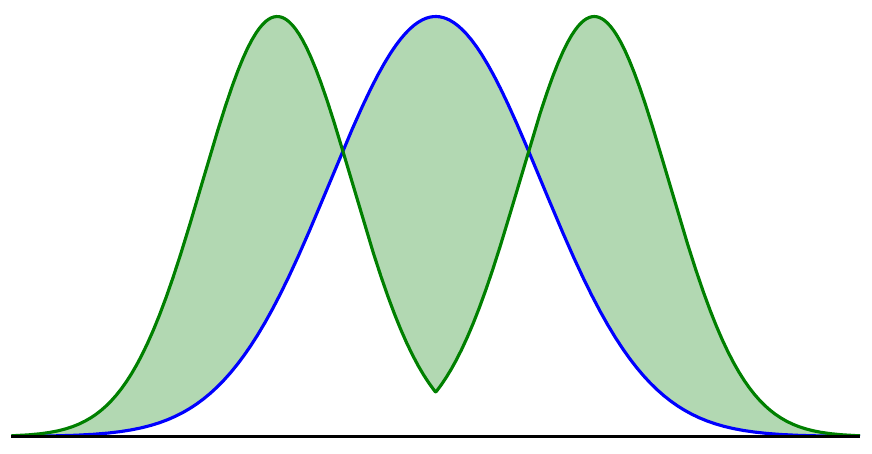}
        no opacity reduction
    \end{minipage}%
    \hfill
    \begin{minipage}{0.42\linewidth}
        \centering
        \includegraphics[width=\linewidth]{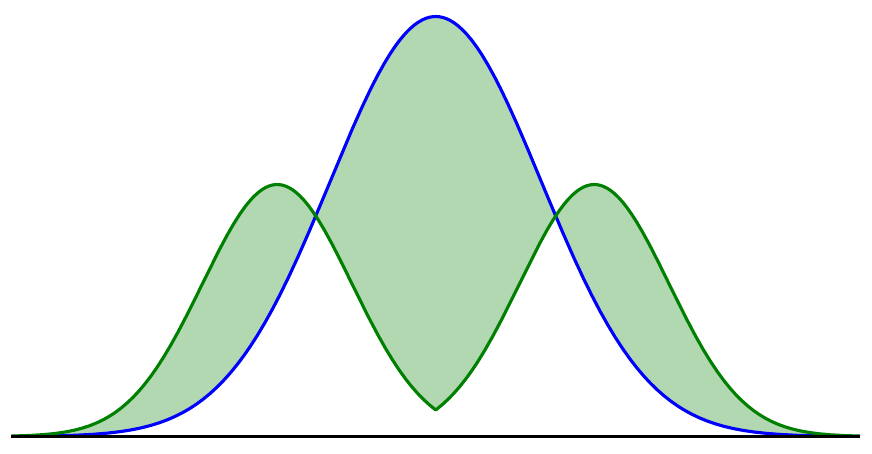}
        opacity reduction
    \end{minipage}
    \caption{After splitting, the density of the corresponding region changes from a unimodal Gaussian distribution (blue curve) to a bimodal Gaussian distribution (green curve).
    If the opacity is not altered after the split (left), the change in the density distribution before and after the split is significant.
    By appropriately reducing the opacity of the split Gaussians (right), the variation in the density distribution can be reduced.
    The green shaded area represents the difference in density distribution before and after splitting.}
    \label{fig:side_by_side}
\end{figure}

\textbf{Opacity}: The split operation transforms the density distribution of the corresponding region from a single-center to a dual-center distribution (see Figure~\ref{fig:side_by_side}).
To reduce the impact of density distribution changes before and after splitting, we lower the opacity of each child Gaussian to $60\%$ of the original Gaussian.
This also alleviates the opacity bias issue that arises when rendering rays sequentially pass through the two child Gaussians.
This adjustment could improve the final rendering quality.
Experimental results can be found in Section~\ref{subsec:ablation}.

\begin{figure}[t]
    \centering
    \includegraphics[width=1.\linewidth]{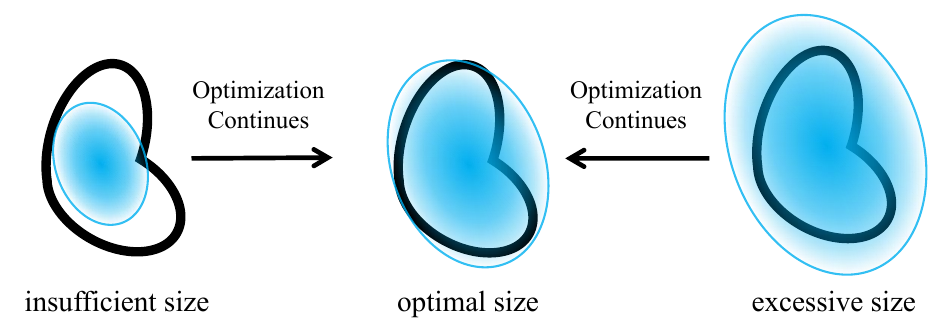}
    \caption{Through the process of backpropagation optimization, the parameters are driven to converge towards a single optimum point. Whether in cases of overfitting or underfitting, the sizes of the Gaussians tend to stabilize at a single consistent value.}
    \label{fig:size}
\end{figure}

\textbf{Split Only}: 3DGS uses clone and split to address under-reconstruction and over-reconstruction.
However, during optimization, the size of Gaussians tends to converge to a value that balances under-reconstruction and over-reconstruction to minimize loss (see Figure~\ref{fig:size}).
Therefore, we only use Long-Axis Split as the densification operation.

\subsection{Recovery-Aware Pruning}
\label{subsec:prune}
3DGS may generate some overfitted Gaussians during the optimization process.
These Gaussians contribute positively in some views but harm the rendering quality in other views.
Overfitted Gaussians may have relatively high opacity, making it difficult to prune them directly using opacity-based methods.

Overfitted Gaussians provide positive contributions in some views and negative contributions in others, while normal Gaussians consistently contribute positively across all views.
After resetting opacity, normal Gaussians will steadily increase their opacity until fully restored, whereas overfitted Gaussians experience fluctuations—opacity increases, decreases, and increases again—resulting in a significantly slower recovery.
This difference in recovery speed can be leveraged for pruning.

The specific operation of Recovery-Aware Pruning involves pruning Gaussians with opacity less than $0.05$ at the $300th$ iteration after resetting opacity, which occurs every $3000$ iterations with a threshold of $0.01$.

The Gaussians pruned by Recovery-Aware Pruning can be divided into two categories.
The first category consists of redundant Gaussians whose opacity is inherently below $0.05$, contributing minimally to rendering.
Pruning these redundant Gaussians during densification reallocates resources to more impactful Gaussians, improving both training performance and generalization.
The second category includes overfitted Gaussians that initially had opacity greater than $0.05$.
These Gaussians make significant contributions in some views, so they may reappear after being pruned.
However, the newly generated Gaussians might no longer overfit. 
After multiple rounds of Recovery-Aware Pruning, the proportion of overfitted Gaussians is greatly reduced, significantly enhancing generalization performance.

While pruning redundant Gaussians improves rendering quality by reallocating resources to higher-contributing Gaussians, Recovery-Aware Pruning achieves its effect by correcting overfitted without altering resource allocation.
A direct consequence is that Recovery-Aware Pruning enhances generalization performance without affecting training performance. Detailed experimental analysis can be found in Section~\ref{subsec:RAP}.

The pruning threshold is only $0.05$, with long intervals and execution limited to the densification phase.
It does not strictly enforce that all Gaussians in the final result must have opacity greater than $0.05$.
In fact, pruning primarily reduces the number of Gaussians with opacity below $0.02$, meaning its impact on rendering semi-transparent objects is limited.

\section{Experiments}
\label{sec:expe}

\begin{table*}[t]
	\centering
	\scalebox{0.54}{ 
		\begin{tabular}{l|cccccc|cccccc|cccccc}
			
			Dataset & \multicolumn{6}{c|}{Mip-NeRF360}  & \multicolumn{6}{c|}{Deep Blending} & \multicolumn{6}{c}{Tanks\&Temples} \\
			Method|Metric
			& $SSIM^\uparrow$   & $PSNR^\uparrow$    & $LPIPS^\downarrow$  & Num/K & Train & FPS 
			& $SSIM^\uparrow$   & $PSNR^\uparrow$    & $LPIPS^\downarrow$  & Num/K & Train & FPS
			& $SSIM^\uparrow$   & $PSNR^\uparrow$    & $LPIPS^\downarrow$  & Num/K & Train & FPS \\
			\hline
			3DGS & 0.815 & 27.48 & 0.216 & 3338 & 21.4 & 169.0 & 0.904 & 29.57 & 0.244 & 2832 & 19.9 & 176.9 & 0.848 & 23.69 & 0.177 & 1847 & 11.7 & 226.8 \\
			EDC-3DGS & 0.815 & 27.52 & 0.221 & 2158 & 16.4 & 304.8 & 0.907 & 29.87 & 0.241 & 1470 & 14.1 & 433.5 & 0.849 & 23.82 & 0.181 & 1074 & 8.3 & 468.9 \\
			\hline
			MiniGS & 0.817 & 27.13 & 0.226 & \colorbox{red!40}{411} & 14.5 & \colorbox{red!40}{680.6} & 0.908 & 30.00 & 0.255 & \colorbox{red!40}{311} & 12.2 & \colorbox{red!40}{1073.4} & 0.835 & 23.30 & 0.205 & \colorbox{red!40}{190} & 7.5 & \colorbox{red!40}{1272.0} \\
			MiniGS-LAS & 0.820 & 27.24 & 0.222 & \colorbox{orange!40}{437} & 15.2 & \colorbox{orange!40}{667.0} & 0.908 & 30.09 & 0.254 & \colorbox{orange!40}{312} & 13.1 & \colorbox{orange!40}{1051.7} & 0.836 & 23.33 & 0.201 & \colorbox{orange!40}{198} & 7.8 & \colorbox{orange!40}{1236.3} \\
			\hline
			TamingGS-S & 0.801 & 27.42 & 0.253 & 689 & \colorbox{orange!40}{5.3} & 300.3 & 0.904 & 29.90 & 0.264 & 375 & \colorbox{orange!40}{3.9} & 433.5 & 0.836 & 23.93 & 0.213 & 300 & \colorbox{orange!40}{3.1} & 504.0 \\
			EDC-TamingGS-S & 0.808 & 27.59 & 0.244 & 689 & \colorbox{red!40}{4.8} & 400.7 & 0.907 & 30.00 & 0.255 & 375 & \colorbox{red!40}{3.5} & 542.1 & 0.846 & 24.18 & 0.199 & 300 & \colorbox{red!40}{2.9} & 626.0 \\
			\hline
			TamingGS-Abs-S & 0.803 & 27.20 & 0.240 & 689 & 6.0 & 250.9 & 0.906 & 29.84 & 0.262 & 375 & 4.8 & 295.1 & 0.837 & 23.77 & 0.210 & 300 & 3.4 & 421.4 \\
			EDC-TamingGS-Abs-S & 0.823 & 27.72 & 0.216 & 689 & 5.4 & 328.1 & \colorbox{orange!40}{0.911} & \colorbox{red!40}{30.14} & 0.247 & 375 & 4.2 & 458.7 & 0.854 & 24.24 & 0.188 & 300 & 3.0 & 596.2 \\
			\hline
			TamingGS-L & 0.820 & 27.89 & 0.215 & 2111 & 10.2 & 196.2 & 0.908 & 30.05 & 0.245 & 1250 & 6.9 & 310.9 & 0.856 & 24.32 & 0.176 & 1000 & 5.7 & 330.7 \\
			EDC-TamingGS-L & \colorbox{orange!40}{0.828} & \colorbox{orange!40}{28.05} & 0.202 & 2111 & 9.5 & 267.2 & 0.909 & \colorbox{orange!40}{30.12} & \colorbox{orange!40}{0.236} & 1250 & 5.8 & 432.7 & \colorbox{orange!40}{0.865} & \colorbox{orange!40}{24.32} & \colorbox{orange!40}{0.157} & 1000 & 5.4 & 429.8 \\
			\hline
			TamingGS-Abs-L & 0.825 & 27.88 & \colorbox{orange!40}{0.197} & 2111 & 10.8 & 175.1 & 0.910 & 29.95 & 0.237 & 1250 & 7.7 & 234.5 & 0.862 & 24.13 & 0.164 & 1000 & 6.3 & 297.7 \\
			EDC-TamingGS-Abs-L & \colorbox{red!40}{0.837} & \colorbox{red!40}{28.15} & \colorbox{red!40}{0.181} & 2111 & 9.9 & 236.4 & \colorbox{red!40}{0.913} & 30.10 & \colorbox{red!40}{0.226} & 1250 & 6.4 & 377.0 & \colorbox{red!40}{0.873} & \colorbox{red!40}{24.48} & \colorbox{red!40}{0.145} & 1000 & 5.4 & 391.3 \\
			\end{tabular}
	}
	\caption{Quantitative results on the Mip-NeRF 360, Deep Blending, and Tanks and Temples datasets. Cells are highlighted as follows: \colorbox{red!40}{best}, and \colorbox{orange!40}{second best}. }
	\label{tab:main}
\end{table*}

\subsection{Datasets and metrics}
\label{subsec:datasets}
We evaluated our method on real-world scenes from the Mip-NeRF 360 \cite{barron2022mip}, Tanks and Temples \cite{knapitsch2017tanks}, and Deep Blending \cite{hedman2018deep} datasets.
We selected all nine scenes from the Mip-NeRF 360 dataset, including five outdoor scenes and four indoor scenes.
For the Tanks and Temples dataset, we chose the train and truck scenes, and for the Deep Blending dataset, we selected the drjohnson and playroom scenes.
As with 3DGS, in each experiment, every 8th image was used as the validation set.
We report peak signal-to-noise ratio (PSNR), structural similarity (SSIM), and perceptual metric (LPIPS) from \cite{zhang2018unreasonable} as quality evaluation metrics.
All three scores were calculated using the methods provided by 3DGS.

\subsection{Implementation}
\label{subsec:impl}
Our code is built upon the open-source 3DGS codebase.
We evaluated the performance improvements of our method on 3DGS \cite{kerbl20233d}, TamingGS \cite{mallick2024taming}, and MiniGS \cite{fang2024mini}.
TamingGS allows for setting an upper limit on the number of Gaussians, and by configuring different thresholds, we tested two versions: Small and Large.
For the Large version, the Gaussian count upper limits were set to $3M$ for five outdoor scenes in the Mip-NeRF 360 dataset, $1.5M$ for drjohnson, and $1M$ for the remaining seven scenes.
For the Small version, the thresholds were set to $1M$, $0.45M$, and $0.3M$, respectively.

We also tested the effect of TamingGS when using the gradient computation method proposed by AbsGS \cite{ye2024absgs}.
For the non-Abs version, all parameters remained consistent with the original implementation.
For the Abs version, the splitting threshold was increased from $0.0002$ to $0.0004$, and the position learning rate combination was changed from $0.00016$–$0.0000016$ to $0.0004$–$0.000002$.
Since EDC can significantly reduce the number of Gaussians under the same splitting threshold (see Section~\ref{subsec:results}), this may result in some scenes failing to reach the upper limit set by TamingGS.
For these scenes, we appropriately reduced the splitting threshold to meet the desired upper limit.
Before the end of the warm-up phase ($300th$ iteration), we performed an opacity pruning operation with a threshold of $0.02$.
This operation alleviates floaters caused by inconsistent lighting in certain scenes.
All results were obtained after $30K$ training iterations.

As mentioned in Section~\ref{subsec:split}, the other axes are shortened to $0.85$ of their original lengths, and the opacity is reduced to $0.6$ of its original value after splitting.
As both values decrease from $1$ to $0$, the difference before and after splitting first decreases and then increases, causing the rendering quality to improve initially and degrade later.
The values $0.85$ and $0.6$ were determined as extrema through testing, and in practice, variations within $\pm 0.05$ yield similar results.

It is worth noting that for methods like MiniGS, which employs two large-scale depth-based initializations during densification and has a very high pruning ratio, both reset opacity and Recovery-Aware Pruning introduce negligible changes (MiniGS does not use reset opacity by default).
Therefore, we only applied Long-Axis Split to MiniGS.

All experiments were conducted on a single 4090D GPU.
All data presented are results we reproduced independently.

\begin{table}[t]
	\centering
	\scalebox{0.6}{
		\begin{tabular}{l|ccccc|cc}
			Dataset & \multicolumn{5}{c|}{Mip-NeRF360}  & \multicolumn{2}{c}{Tanks\&Temples}\\
			Method|Scene & bicycle & flowers & garden & stump & treehill & train & truck \\
			\hline 
			3DGS & 25.213 & 21.539 & 27.361 & 26.539 & 22.495 & 21.958 & 25.414 \\
			EDC-3DGS & 25.267 & 21.630 & 27.481 & 26.661 & 22.614 & 22.035 & 25.598 \\
			\hline 
			MiniGS & 25.202 & 21.403 & 26.863 & 27.125 & 22.699 & 21.575 & 25.027 \\
			MiniGS-LAS & 25.275 & 21.622 & 27.055 & 27.231 & 22.717 & 21.572 & 25.082 \\
			\hline 
			TamingGS-S & 24.934 & 21.562 & 27.293 & 26.433 & 23.059 & 22.446 & 25.424 \\
			EDC-TamingGS-S & 25.210 & 21.634 & 27.509 & 26.713 & 23.011 & 22.592 & 25.763 \\
			\hline 
			TamingGS-Abs-S & 25.109 & 21.183 & 27.095 & 26.394 & 22.775 & 22.518 & 25.019 \\
			EDC-TamingGS-Abs-S & 25.597 & 21.684 & 27.555 & 27.016 & 23.017 & 22.714 & 25.763 \\
			\hline 
			TamingGS-L & 25.417 & 21.842 & 27.748 & 26.717 & 23.027 & 22.689 & 25.946 \\
			EDC-TamingGS-L & 25.580 & 22.147 & 27.955 & 26.833 & 23.035 & 22.485 & 26.150 \\
			\hline 
			TamingGS-Abs-L & 25.550 & 21.398 & 27.735 & 26.834 & 22.591 & 22.483 & 25.768 \\
			EDC-TamingGS-Abs-L & 25.863 & 21.770 & 28.085 & 27.213 & 22.853 & 22.580 & 26.387 
		\end{tabular}
	}
	\caption{PSNR scores for outdoor scenes from the Mip-NeRF 360 and Tanks and Temples datasets.}
	\label{tab:psnr_out}
\end{table}

\begin{table}[t]
	\centering
	\scalebox{0.62}{
		\begin{tabular}{l|cccc|cc}
			Dataset & \multicolumn{4}{c|}{Mip-NeRF360}  & \multicolumn{2}{c}{Deep Blending}\\
			Method|Scene & bonsai & counter & kitchen & room & drjohnson & playroom \\
			\hline 
			3DGS & 32.242 & 29.016 & 31.474 & 31.446 & 29.119 & 30.019 \\
			EDC-3DGS & 32.039 & 28.896 & 31.470 & 31.650 & 29.485 & 30.264 \\
			\hline 
			MiniGS & 30.966 & 28.306 & 30.774 & 30.864 & 29.501 & 30.501 \\
			MiniGS-LAS & 31.032 & 28.359 & 30.769 & 31.132 & 29.542 & 30.646 \\
			\hline 
			TamingGS-S & 31.889 & 29.064 & 30.885 & 31.673 & 29.516 & 30.283 \\
			EDC-TamingGS-S & 31.947 & 29.142 & 31.451 & 31.714 & 29.628 & 30.378 \\
			\hline 
			TamingGS-Abs-S & 31.559 & 29.036 & 30.382 & 31.281 & 29.396 & 30.290 \\
			EDC-TamingGS-Abs-S & 31.986 & 29.351 & 31.529 & 31.741 & 29.673 & 30.606 \\
			\hline 
			TamingGS-L & 32.761 & 29.469 & 31.973 & 32.074 & 29.628 & 30.466 \\
			EDC-TamingGS-L & 32.806 & 29.640 & 32.315 & 32.183 & 29.634 & 30.610 \\
			\hline 
			TamingGS-Abs-L & 32.726 & 29.562 & 32.277 & 32.238 & 29.511 & 30.386 \\
			EDC-TamingGS-Abs-L & 32.908 & 29.836 & 32.501 & 32.279 & 29.735 & 30.466 \\
		\end{tabular}
	}
	\caption{PSNR scores for indoor scenes from the Mip-NeRF 360 and Deep Blending datasets.}
	\label{tab:psnr_in}
\end{table}

\subsection{Quantitative Analysis}
\label{subsec:results}

We selected $6$ methods: 3DGS, MiniGS, TamingGS-S, TamingGS-L, TamingGS-Abs-S, and TamingGS-Abs-L to evaluate the improvements brought by EDC.
As shown in Table~\ref{tab:main}, our method significantly enhances the quality of each benchmark.
The PSNR scores for each scene can be found in Table~\ref{tab:psnr_in} and Table~\ref{tab:psnr_out}.

EDC-TamingGS-Abs-L achieves the best rendering quality while using a significantly smaller number of Gaussians compared to 3DGS.
Even under resource-constrained conditions, EDC-TamingGS-Abs-S achieves rendering quality that is notably superior to 3DGS.
In the following sections, we will analyze the results of each set of experiments.

\textbf{3DGS}: In EDC-3DGS, we maintained the same parameters as 3DGS, but the number of Gaussians differs substantially.
One reason for this is that with Recovery-Aware Pruning, some regions undergo a process of pruning, regenerating, and re-pruning Gaussians, which impacts the densification speed.
On the other hand, EDC achieves the same rendering quality with fewer Gaussians, meaning that under the same splitting threshold, the upper limit of Gaussians that EDC can achieve is lower.
Even with fewer Gaussians, EDC still improves the rendering quality of 3DGS.

\begin{figure*}[t]
    \centering
    \includegraphics[width=1.\linewidth]{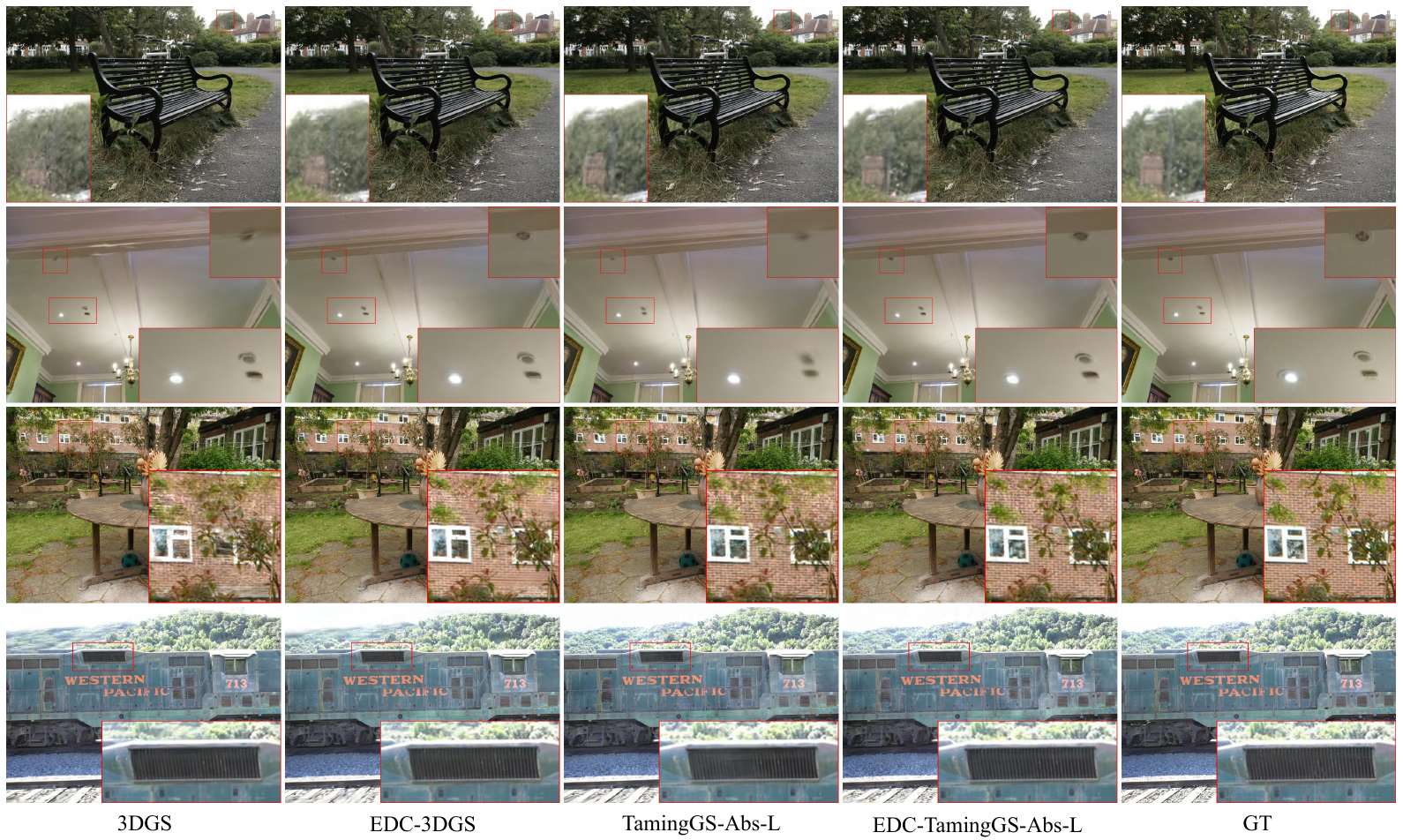}
    \caption{Qualitative analysis of the improvements brought by our approach across multiple scenes.}
	\label{fig:main}
	\vspace{-1em}
\end{figure*}

\textbf{MiniGS}: MiniGS employs two large-scale depth reinitializations, making its densification process no longer fully dependent on ADC.
However, Long-Axis Split still enhances the rendering quality of MiniGS without introducing significant overhead.
This demonstrates that our method can also be applied to compression-related tasks.

\textbf{TamingGS}: TamingGS previously achieved state-of-the-art performance in improving when to densify.
Our tests show that incorporating the improvements proposed by AbsGS can further enhance the performance of TamingGS.
Since TamingGS manually constrains the growth curve of Gaussian numbers, we are able to eliminate the influence of the number of Gaussians on the results and focus on analyzing the quality improvement brought by EDC.

In all four TamingGS experimental groups, our method brings significant quality improvements.
Notably, EDC-TamingGS-Abs-L achieves a PSNR score of $28.15$ on the Mip-NeRF 360 dataset.
With the same number of Gaussians, our method also achieves over a $30\%$ increase in rendering speed.
This is because our method avoids overlapping Gaussians after splitting, reducing the average number of Gaussians traversed by rendering rays per pixel.

The peak number of Gaussians during the training of MiniGS is approximately ten times the final count, while for TamingGS, the peak equals the final count, indicating that TamingGS-S has a lower training requirements.
After applying our method, only one-fifth of the training requirements is needed to achieve rendering quality surpassing 3DGS, which facilitates the wider adoption of 3DGS.

\subsection{Qualitative Analysis}
\label{subsec:results2}

In Figure~\ref{fig:main}, we present the improvements of our method compared to 3DGS and TamingGS-Abs-L, with key areas magnified.
Thanks to more precise splitting, our method is able to better recover details, such as the distant streetlights in the bicycle scene, the ceiling lights and smoke detectors in the drjohnson scene.
Our method also performs better in regions with dense lines, such as the distant brick wall in the garden scene and the radiator in the train scene.

\subsection{Ablation Experiments}
\label{subsec:ablation}

We tested the impact of each improvement (including reduce the opacity of child Gaussians) on the Mip-NeRF 360 dataset  (see Table~\ref{tab:ablation}).
We also tested the results on two other datasets, but due to their higher variability, we only used Mip-NeRF 360 for the quantitative analysis to ensure experimental rigor.
The baseline used for testing was TamingGS-Abs-S.

\begin{figure}[t]
    \centering
    \includegraphics[width=0.7\linewidth]{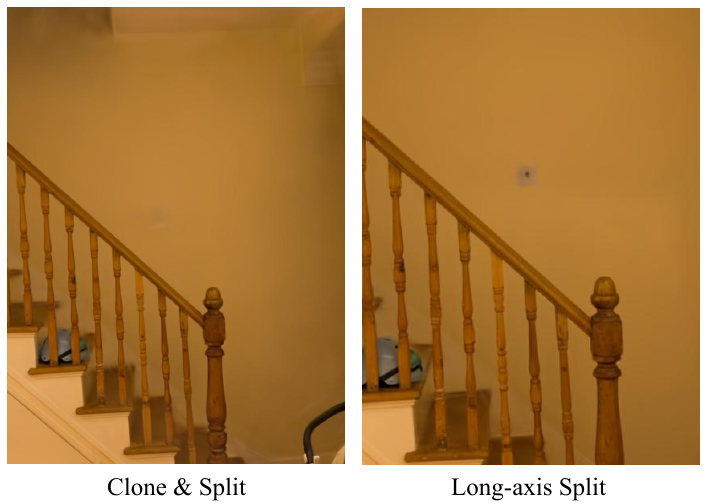}
    \caption{Qualitative comparison of two densification operations in the playroom scene, using TamingGS-Abs-S.}
	\label{fig:two}
\end{figure}

\textbf{Long-Axis Split}: When applying only Long-Axis Split, the average PSNR improved by $0.27$.
This demonstrates that Long-Axis Split is indeed more efficient than clone and split.
Long-Axis Split minimizes the differences before and after splitting, leading to better fitting of details.
As shown in Figure~\ref{fig:two}, when Long-Axis Split is not applied, the switch on the wall is barely reconstructed correctly.
However, after applying Long-Axis Split, the switch becomes clearly visible.
Additionally, reducing opacity after splitting can also provide a certain performance boost.

\textbf{Recovery-Aware Pruning}: Applying only Recovery-Aware Pruning also leads to significant quality improvements.
This is partly due to the elimination of redundancy and partly due to the mitigation of overfitting.
Notably, when Long-Axis Split is used, Recovery-Aware Pruning can lead to even more substantial improvements in PSNR scores.
Long-Axis Split uses a fixed splitting dimension, which reduces training fluctuations caused by probabilistic sampling but increases the likelihood of overfitting.
Moreover, since Long-Axis Split has a stronger ability to fit detailed regions, these areas are more prone to overfitting.
Recovery-Aware Pruning helps reduce the impact of overfitting and is highly complementary to Long-Axis Split.
Therefore, we recommend applying both methods simultaneously in most cases.

\begin{table}[t]
	\centering
	\scalebox{0.75}{
		\begin{tabular}{l|ccc}
			
			Method|Metric & $SSIM^\uparrow$ & $PSNR^\uparrow$ & $LPIPS^\downarrow$ \\
			\hline 
			Base & 0.803 & 27.20 & 0.240 \\
			+Long-Axis Split & 0.812 & 24.47 & 0.233 \\
			+Recovery-Aware Pruning & 0.812 & 27.41 & 0.227 \\
			\hline
			No opacity Reduction & 0.821 & 27.67 & 0.217 \\
			Full & 0.823 & 27.72 & 0.216 \\

		\end{tabular}
	}
	\caption{Results of the ablation study on the Mip-NeRF 360 dataset.}
	\label{tab:ablation}
	\vspace{-1em}
\end{table}

\subsection{Effectiveness of Recovery-Aware Pruning}
\label{subsec:RAP}
The experimental setup in this subsection is based on TamingGS-Abs + Long-Axis Split/ TamingGS-Abs, with a Gaussian cap of $1M$ and the scene being bicycle.

We tested the results without pruning, pruning Gaussians with opacity less than 0.05 before resetting opacity (pruning redundant Gaussians), and using Recovery-Aware Pruning.
The results are shown in Table~\ref{tab:prune}.
Firstly, pruning redundant Gaussians improves both training performance and generalization performance.
Secondly, compared to pruning redundant Gaussians, Recovery-Aware Pruning significantly enhances generalization performance but does not notably improve training performance, consistent with the analysis in Section~\ref{subsec:prune}.
Pruning only drastically reduces the proportion of Gaussians with opacity less than $0.02$, whose rendering contribution is minimal.

\begin{table}[t]
	\centering
	\scalebox{0.75}{
		\begin{tabular}{l|cc|ccc}
			
			Method|Metric & Test & Train & $o<0.02$ & $o<0.05$ & $o<0.1$ \\
			\hline 
			LAS-Base & 25.108 & 24.642 & 15.378 & 10.298 & 12.139 \\
			LAS-Prune 0.05 & 25.190 & 24.788 & 3.809 & 5.836 & 14.201 \\
			LAS-RAP & 25.294 & 24.811 & 0.910 & 4.737 & 13.181 \\
			\hline 
			Base & 25.257 & 25.128 & 20.492 & 10.687 & 10.573 \\
			Prune 0.05 & 25.420 & 25.373 & 5.267 & 6.367 & 14.243 \\
			RAP & 25.597 & 25.368 & 0.971 & 4.166 & 11.893 \\

		\end{tabular}
	}
	\caption{The analysis results of Recovery-Aware Pruning are presented as follows: the first three rows represent the results when using Long-Axis Split, and the last three rows represent the results when using Clone \& Split. The testing scenario used was bicycle.}
	\label{tab:prune}
	\vspace{-1em}
\end{table}

\subsection{Analyze of Clone and Reset Opacity}
\label{subsec:reset}
In Section~\ref{subsec:limitations}, we proposed that the clone operation relies on the current gradient to separate the cloned Gaussians.
However, this method does not always work effectively.
To visually demonstrate the impact of Gaussian overlap, we moved the cloned Gaussians along the major axis of the parent by $1/4$ of its axis length and compared the results with the original clone operation.
The comparison was conducted using TamingGS-Abs, with a Gaussian cap of 1M, and the scene being bicycle.
The PSNR values for using clone + split, adjusted clone + split, and fully using split are $24.929$, $24.989$, and $25.047$, respectively.
Avoiding complete overlap between Gaussians improves rendering quality, however using only split achieved the best results.

We also conducted a simple test on the impact of resetting opacity itself using TamingGS-Abs-S + Long-Axis Split on the bicycle scene.
The PSNR was $25.245$ when using reset opacity and $25.206$ when not using it.
Additionally, reset opacity reduced the proportion of Gaussians with opacity greater than $0.9$ from $0.12$ to $0.03$, with the excess allocation evenly distributed to Gaussians with opacity less than $0.5$.
This may suggest that a more balanced distribution of rendering contributions is beneficial for high-quality reconstruction.

\section{Conclusion}
\label{sec:clu}
3DGS employs Adaptive Density Control (ADC) to increase the number of Gaussians.
The densification process is interleaved with optimization, and to improve efficiency, we should minimize the negative impact caused by densification.
However, cloning and splitting operations fail to meet this requirement.
To address this, we propose a more precise method called Long-Axis Splitting, which minimizes the differences before and after splitting.
Additionally, our proposed Recovery-Aware Pruning successfully prunes overfitted Gaussians by leveraging the difference in recovery speed after resetting opacity.
Overall, our method achieves state-of-the-art performance in rendering quality.

{
    \small
    \bibliographystyle{ieeenat_fullname}
    \bibliography{main}
}


\end{document}